\documentclass[10pt,journal]{IEEEtran}
\ifCLASSOPTIONcompsoc
\else
\fi

\ifCLASSINFOpdf
\else
\fi

\usepackage[numbers,sort&compress]{natbib}
\usepackage{graphicx}
\usepackage{amsmath,amsfonts,amssymb,amsthm} 
\usepackage{algorithm}
\usepackage{algorithmic}
\usepackage{subfigure}
\renewcommand{\Re}{\mathbb{R}}
\newcommand{\Id}{\mathtt{I}}

\newtheorem{problem}{Problem}

\begin{document}

\title{Robust Subspace Clustering with\\ Compressed Data}

\author{Guangcan Liu,~\IEEEmembership{Senior Member,~IEEE}, Zhao Zhang, ~\IEEEmembership{Senior Member,~IEEE}, Qingshan Liu,~\IEEEmembership{Senior Member,~IEEE}, and Hongkai Xiong,~\IEEEmembership{Senior Member,~IEEE}
\IEEEcompsocitemizethanks{\IEEEcompsocthanksitem G. Liu is with B-DAT and CICAEET, School of Automation, Nanjing University of
Information Science and Technology, Nanjing, China 210044. Email: gcliu@nuist.edu.cn.\protect
\IEEEcompsocthanksitem Z. Zhang is with School of Computer Science and Technology, Soochow University, Suzhou, China 215006. Email: cszzhang@gmail.com.\protect
\IEEEcompsocthanksitem Q. Liu is with B-DAT, School of Automation, Nanjing University of
Information Science and Technology, Nanjing, China 210044. Email: qsliu@nuist.edu.cn.\protect
\IEEEcompsocthanksitem H. Xiong is with Department of Electronic Engineering, Shanghai Jiao Tong University, 800 Dongchuan Road, Shanghai, China 200240. E-mail: xionghongkai@sjtu.edu.cn.
}
\thanks{}}
\maketitle

\begin{abstract} Dimension reduction is widely regarded as an effective way for decreasing the computation, storage and communication loads of data-driven intelligent systems, leading to a growing demand for statistical methods that allow analysis (e.g., clustering) of compressed data. We therefore study in this paper a novel problem called \emph{compressive robust subspace clustering}, which is to perform robust subspace clustering with the compressed data, and which is generated by projecting the original high-dimensional data onto a lower-dimensional subspace chosen at random. Given only the compressed data and sensing matrix, the proposed method, \emph{row space pursuit} (RSP), recovers the authentic row space that gives correct clustering results under certain conditions. Extensive experiments show that RSP is distinctly better than the competing methods, in terms of both clustering accuracy and computational efficiency.
\end{abstract}

\section{Introduction}\label{sec:introduction}
\IEEEPARstart{A}{long} with the evolvement of data collection technology, the dimension of data is now getting higher and higher. For example, one can easily use the latest camera phones to take high-quality photos in a resolution of 40 or more megapixels. In general, when the data dimension increases, the cost in storing, transmitting and analyzing data will inevitably rise. What is more, the increase in data dimension is actually much faster than the advance in communication, storage and computation power. As a consequence, it is often desirable to reduce the dimensionality of data. However, given that the data of interest has been compressed via dimension reduction, a natural question to ask is \emph{how to analyze the structure of the original data by only accessing the compressed data}.

To investigate the highlighted problem, one essentially needs to integrate dimension reduction and pattern analysis into a unified framework. Among various dimension reduction methods, we would like to consider the well-known Random Projection (RP)~\cite{rp:kdd:2001}, in which the original $m$-dimensional data points are projected to a $p$-dimensional ($p\ll{}m$) subspace, using some sensing matrix $R\in\mathbb{R}^{p\times{}m}$ generated at random. Unlike the data-dependent methods such as Principal Component Analysis (PCA) and Principal Component Pursuit (PCP)~\cite{Candes:2011:JournalACM}, which must be trained before being applied to the data, RP needs no training procedure and is therefore computationally efficient. Even more, the sensing matrix $R$ is randomly generated and thus could be shared across multiple devices without transmission. Due to these advantages, RP is often the most preferable choice for reducing the dimension of data stored on resource-constrained devices, e.g., satellite-borne sensors~\cite{cppca:2009}. Regarding the pattern analysis problems, \emph{robust subspace clustering}~\cite{vidal:spm:2011,icml_2010_lrr,tpami_2013_lrr,robust_spc}, the task of grouping together the data points lying approximately on the same (linear) subspace, has considerable practical and scientific significance. In fact, as pointed out by~\cite{candes:2012:sr:outliers}, robust subspace clustering is a representative setting of unsupervised learning. So, in the spirit of unifying data compression with pattern analysis, we would suggest considering the following problem that combines RP with robust subspace clustering.

\begin{problem}[Compressive Robust Subspace Clustering]~\label{pb:csc} Let $X=[x_1,\cdots,x_n]\in\mathbb{R}^{m\times{}n}$ store a collection of $n$ $m$-dimensional points approximately drawn from a union of $k$ subspaces. Suppose that $R\in\mathbb{R}^{p\times{}m} (p\ll{}m)$ is a random Gaussian matrix whose columns have unit lengths. Denote $M\triangleq{}RX\in\mathbb{R}^{p\times{}n}$. Given the compressed data $M$ and the sensing matrix $R$, the goal is to segment all points into their respective subspaces.
\end{problem}

Due to its significance in science and application, robust subspace clustering has received extensive attention in the literature, e.g.,~\cite{cacm_1981_ransac,cvpr_2003_ksubspace,cvpr_2004_em,corr_2009_kflats,ijcv_1998_factor,siam_2008_gpca,tpami_2008_acl,cvpr_2009_ssc,Chen:2009:SCC,icml_2010_lrr,tpami_2013_lrr,yuxiang:nips:2013,talalkar:iccv:2013,liu:2011:iccv,Lu:2012:RES,robust_spc,paolo:cvpr:2011,Peng2013Scalable,zhang2014fast,liu:tpami:2016,liu:tpami:2017,Peng:2016:DSC,rah:icml:2017,LU:pami:2018,tyb:2017:peng}. However, most existing methods are specific to the original uncompressed data $X$, and we have spotted only sparse researches relevant to the setup of Problem~\ref{pb:csc}. Namely, in~\cite{mao:2014:compressed,ruta:2012:compressed}, it is shown that the subspace principal angles before and after RP compression are mostly unchanged, which means that one may simply input $M$ into some existing subspace clustering methods such as Shape Interaction Matrix (SIM)~\cite{ijcv_1998_factor}, Sparse Subspace Clustering (SSC)~\cite{cvpr_2009_ssc,ssc:2013:pami} and Low-Rank Representation (LRR)~\cite{tpami_2013_lrr,liu:tpami:2016}. This, however, may not work for Problem~\ref{pb:csc}, in which the original data $X$ could be contaminated by gross errors; namely,
\begin{align*}
X = L_0 + S_0,
\end{align*}
where $L_0$ stores the authentic samples lying exactly on the subspaces and $S_0$ corresponds to the possible errors. In the presence of gross errors, i.e., the entries in $S_0$ have large magnitudes, it will lead to very poor results by simply applying subspace clustering methods to the compressed data $M$. This is because the projection procedure could change the statistical properties of the errors. For example, consider the case where the gross errors are entry-wisely sparse; that is, a small fraction of the entries in $S_0$ are nonzero and have large magnitudes. In the compressed data matrix $M$, however, the errors may spread to every entries of the matrix, thereby $RS_0$ is often a dense matrix with large values. The resulted problem---segmenting the points in $M$ into correct subspaces purely based on $M$ which itself is corrupted by dense gross errors---is indeed intractable.

In order to study Problem~\ref{pb:csc} under the context of sparse errors, we propose a simple yet effective method termed \emph{row space pursuit} (RSP). Given the compressed data matrix $M$ and sensing matrix $R$, RSP recovers not only the row space of the clean data (i.e., $L_0$) but also the possible gross errors. Since the authentic row space (i.e., row space of $L_0$) provably determines the true subspace membership of the data points, the final clustering results are obtained by simply using the recovered row space as input to perform K-Means clustering. In general, RSP owns a computational complexity of only $O(mnp)$ and can therefore fast segment a large number of high-dimensional data points. What is more, most of the computational resources required by RSP are spent on matrix multiplications, which are easy to accelerate by parallel algorithms. Extensive experiments on high-dimensional and large-scale datasets demonstrate the superior performance of RSP, in terms of both clustering accuracy and computational efficiency. In effect, RSP can even maintain comparable accuracies to the prevalent methods that perform clustering using the original high-dimensional, uncompressed data.
\section{Problem Formulation and Analysis}\label{sec:formulation}
Formally, the regime underlying a collection of points approximately drawn from a union of $k$ subspaces could be modeled as $X = L_0 + S_0$, where $L_0$ and $S_0$ correspond to the components of the authentic samples and possible errors, respectively. The word ``error'', in general, refers to the deviation between the model assumption (i.e., subspaces) and the observed data. In practice, the errors could exhibit as white noise~\cite{CandesPIEEE}, missing entries~\cite{liu:nips:2017}, outliers~\cite{jmlr_2012_lrr,liu:tpami:2016} and corruptions~\cite{Candes:2011:JournalACM}. In this paper, we would like to focus on the setting of gross corruptions studied in PCP~\cite{Candes:2011:JournalACM}; namely, $S_0$ is entry-wisely sparse and the values in $S_0$ are arbitrarily large.

As shown in~\cite{ijcv_1998_factor,tpami_2013_lrr,yuxiang:nips:2013}, the row space of $L_0$ can lead to exact subspace clustering under certain conditions. Hence, Problem~\ref{pb:csc} would be mathematically formulated as a problem called \emph{compressive row space recovery}:
\begin{problem}[Compressive Row Space Recovery]~\label{pb:crsr} Let $L_0\in\mathbb{R}^{m\times{}n}$ with (skinny) SVD $U_0\Sigma_0V_0^T$ and rank $r_0$ store a set of $n$ $m$-dimensional authentic samples strictly drawn from a union of $k$ subspaces, where $V_0\in\mathbb{R}^{n\times{}r_0}$. Let $R\in\mathbb{R}^{p\times{}m} (r_0< p \ll m)$ be a random Gaussian matrix whose columns have unit $\ell_2$ norms. Suppose that the data matrix $X$ is generated by $X = L_0+S_0$, with $S_0$ being an entry-wisely sparse matrix corresponding to the possible errors. Denote by $M\triangleq{}RX\in\mathbb{R}^{p\times{}n}$ the compressed data matrix. Given $M$ and $R$, the goal is to identify $V_0V_0^T$ and $S_0$.
\end{problem}

The above problem is essentially a generalization of the \emph{subspace recovery} problem studied in~\cite{tpami_2013_lrr}. To approach Problem~\ref{pb:crsr}, one may consider Compressive Sparse Matrix Recovery (CSMR)~\cite{mardani:2013:tit}, which is a variation of Compressive Principal Component Pursuit (CPCP)~\cite{john:2012:cpcp}. Given $M$ and $R$, CSMR strives to recover $RL_0$ and $S_0$ by solving the following convex optimization problem:
\begin{align}\label{eq:csmr}
\min_{A\in\mathbb{R}^{p\times{}n},S\in\mathbb{R}^{m\times{}n}} \|A\|_* + \lambda \|S\|_1, \textrm{ s.t. }M = A + RS,
\end{align}
where $\|\cdot\|_1$ denotes the $\ell_1$ norm of a matrix seen as a long vector. Under certain conditions, it is provable that CSMR strictly succeeds in recovering both $RL_0$ and $S_0$. However, as clarified in~\cite{mardani:2013:tit}, CSMR is actually designed for the case where $m\gg{}n\gg{}r_0$, i.e., $L_0$ is a tall, low-rank matrix such that $RL_0$ is still low rank. In the cases of square or fat matrices, the recovery ability of CSMR is quite limited, because in this case $RL_0$ could be high rank or even full rank. To achieve better results, we shall propose a new method termed RSP.

\section{Compressive Robust Subspace Clustering via Row Space Pursuit}\label{sec:main}
In this section, we shall detail the proposed RSP method for compressive robust subspace clustering.
\subsection{Compressive Row Space Recovery by RSP}\label{sec:main:rsp} The formula of RSP is derived as follows. Denote by $U_0\Sigma_0V_0^T$ and $r_0$ the SVD and rank of $L_0$, respectively. Since $M = R(L_0+S_0)$, we could construct a matrix $P\in\Re^{n\times{}n}$ to annihilate $L_0$ on the right, i.e., $L_0P = 0$. This can be easily done by taking $P = \Id - V_0V_0^T$, with $\Id$ being the identify matrix. That is,
\begin{align}\label{eq:constraint}
(M - RS_0)(\Id - V_0V_0^T) = RL_0(\Id - V_0V_0^T)=0.
\end{align}
Hence, we may seek both $V_0$ and $S_0$ by the following non-convex program:
\begin{align}\label{eq:rsp:noiseless}
&\min_{V\in\mathbb{R}^{n\times{}r},S\in\mathbb{R}^{m\times{}n}} \|S\|_1, \\\nonumber
&\textrm{ s.t. } (M - RS)(\Id - VV^T) = 0, V^TV = \Id,
\end{align}
where $r$ ($r_0\leq{}r<p$) is taken as a parameter. In order to attain an exact recovery to the authentic row space $V_0$, we would need $r=r_0$. Yet, to obtain superior clustering results in practice, exact recovery is not indispensable, and it is indeed unnecessary for the parameter $r$ to strictly equal to the true rank $r_0$, as will be shown in our experiments. There is also an intuitive explanation for this phenomenon. That is, the equality in \eqref{eq:constraint} always holds when $V_0$ is replaced by any other space that includes $V_0$ as a subspace.

\textbf{Analysis:} We shall briefly analyze the performance of the RSP program~\eqref{eq:rsp:noiseless}, under the context of Problem~\ref{pb:crsr}. To do this, we first consider an equivalent version of~\eqref{eq:rsp:noiseless}:
\begin{align}\label{eq:rsp:equi}
\min_{P,S} \|S\|_1, \textrm{ s.t. } (M - RS)(\Id - P) = 0, P\in\Xi,
\end{align}
where $\Xi = \{VV^T: V\in\mathbb{R}^{n\times{}r}, V^TV=\Id\}$ is the set of orthogonal projections onto a $r$-dimensional subspace. For the sake of simplicity, assume that $r=r_0$. Whenever $S = S_0$, it is provable that $P=V_0V_0^T$ is the only feasible solution to the problem in~\eqref{eq:rsp:equi}. More precisely, provided that $p\geq{}r_0$, it is almost surely (i.e., with probability 1) that the row space of $RL_0$ is exactly $V_0$~\cite{siam:2011:halko}. On the other hand, given $P=V_0V_0^T$, the problem in~\eqref{eq:rsp:equi} turns into a \emph{sparse signal recovery} problem explored in~\cite{candes:2005:tit}; namely,
\begin{align}\label{eq:rsp:con}
\min_{y} \|y\|_1, \textrm{ s.t. } b = \Phi{}y,
\end{align}
where $\Phi=(\Id - V_0V_0^T)\otimes{}R$ and $b=\mathrm{vec}(M(\Id - V_0V_0^T))$. Here, the symbols $\otimes$ and $\mathrm{vec}(\cdot)$ denote the Kronecker product and the vectorization of a matrix into a long vector, respectively. Since $\Id - V_0V_0^T$ is an orthogonal projection, $\Phi$ may still satisfy the so-called Restricted Isometry Property (RIP)~\cite{candes:2005:tit}. As a result, according to~\cite{candes:2005:tit}, the convex program in~\eqref{eq:rsp:con} may identify $\mathrm{vec}(S_0)$ with overwhelming probability, as long as $p\geq{}c\|S_0\|_0/n$ for some numerical constant $c$, where $\|\cdot\|_0$ is the $\ell_0$ pseudo-norm of a matrix, i.e., the number of nonzero entries of a matrix.

In summary, the results in~\cite{candes:2005:tit,siam:2011:halko} have already proven that $(P=V_0V_0^T, S=S_0)$ is a critical point to the non-convex problem in~\eqref{eq:rsp:equi}. However, due to the orthonormal constraint $V^TV=\Id$, it would be hard to obtain a stronger guarantee. Thus, in this paper we would like to focus on the empirical performance of RSP. Still, the above analysis provides some useful clues for understanding the behaviors of RSP. Namely, to obtain exact or near exact recoveries to $V_0V_0^T$ and $S_0$, the number of random projections $p$ has to obey the following two conditions:
\begin{align}\label{eq:condition}
p\geq{}r_0\textrm{ and } p\geq{}c\|S_0\|_0/n.
\end{align}
For convenience, hereafter, we shall consistently refer to the quantity $\|S_0\|_0/n$ as the \emph{corruption size}.

\subsection{Optimization Algorithm}\label{sec:main:optimization}
The observed data in reality is often contaminated by noise, and thus we shall consider instead the following non-convex program that can also approximately
solve the problem in~\eqref{eq:rsp:noiseless}:
\begin{align}\label{eq:rsp:noisy}
&\min_{V\in\mathbb{R}^{n\times{}r},S\in\mathbb{R}^{m\times{}n}} \lambda\|S\|_1 +\\\nonumber
&\frac{1}{2}\|(M - RS)(\Id - VV^T)\|_F^2, \textrm{ s.t. } V^TV = \Id,
\end{align}
where $\|\cdot\|_F$ denotes the Frobenius norm of a matrix and $\lambda>0$ is a parameter.

Although non-convex as a whole, the problem in~\eqref{eq:rsp:noisy} is indeed easy to solve while one of $V$ and $S$ is given, thereby it is suitable to solve~\eqref{eq:rsp:noisy} by the first-order methods established in the literature~\cite{lin_alm,Lu:pami:2018b,proximal:2009:mp}. We choose to use the alternating proximal method established in~\cite{proximal:2009:mp}. Let $(V_t,S_t)$ be the solution estimated at the $t$th iteration. Denote
\begin{align*}
g(V,S)\triangleq\frac{1}{2}\|(M - RS)(\Id - VV^T)\|_F^2.
\end{align*}
Then the solution to~\eqref{eq:rsp:noisy} is updated via iterating the following two procedures:
\begin{align}\label{eq:proximal}
&V_{t+1}=\arg\min_{V} g(V,S_t), \textrm{ s.t. }V^TV=\Id,\\\nonumber
&S_{t+1}=\arg\min_{S} \frac{\lambda}{\rho}\|S\|_1+\frac{1}{2}\|S - (S_t-\frac{\partial_S{}g(V_{t+1},S)}{\rho})\|_F^2,
\end{align}
where $\rho>0$ is a penalty parameter and $\partial_S{}g(V_{t+1},S)$ is the partial derivative of $g(V,S)$ with respect to the variable $S$ at $V=V_{t+1}$; namely,
\begin{align}\label{eq:gradient}
&\partial_S{}g(V_{t+1},S) = R^T(RS-M)(I-V_{t+1}V_{t+1}^T)\\\nonumber
&=R^T(RS-RSV_{t+1}V_{t+1}^T - M + MV_{t+1}V_{t+1}^T).
\end{align}
According to~\cite{proximal:2009:mp}, the penalty parameter could be set as $\rho = 1.1\|R\|^2$, where $\|\cdot\|$ is the operator norm of a matrix, i.e., the largest singular value.

The two optimization problems in~\eqref{eq:proximal} both have closed-form solutions. More precisely, the $V$-subproblem is solved by finding the top $r$ eigenvectors of a semi-positive definite matrix, $(M-RS_t)^T(M-RS_t)$. To do this, one actually just needs to calculate the top $r$ right singular vectors of $M - RS_t$, which is a $p\times{}n$ matrix. The solution to the $S$-subproblem is given by
\begin{align}\label{eq:s-sub}
S_{t+1}= \mathcal{H}_{\lambda/\rho}[S_t-\frac{\partial_S{}g(V_{t+1},S)}{\rho}],
\end{align}
where $\mathcal{H}_{\lambda/\rho}[\cdot]$ denotes the entry-wise shrinkage operator with parameter $\lambda/\rho$. The whole optimization procedure is also summarized in Algorithm~\ref{alg:ap}.
\begin{algorithm}[htb]
\caption{Solving the problem in~\eqref{eq:rsp:noisy} by the alternating proximal method}
\label{alg:ap}
\begin{algorithmic}[1]
\STATE \textbf{input}: $M$, $R$, $r$ and $\lambda$.
\STATE \textbf{parameters}: $\rho=1.1\|R\|^2$
\STATE \textbf{Output}: $V$ and $S$.
\STATE \textbf{Initialization}: $S = 0$.
\REPEAT
\STATE compute the matrix $M-RS$.
\STATE update $V$ using the top $r$ right singular vectors of $M-RS$.
\STATE compute the gradient given in~\eqref{eq:gradient}.
\STATE update $S$ by~\eqref{eq:s-sub}.
\UNTIL{convergence}
\end{algorithmic}
\end{algorithm}

Empirically, the convergence of Algorithm~\ref{alg:ap} is determined when the objectives of two adjacent iterations differ by no more than $10^{-9}\|M\|_F^2$; namely, $|o_{t+1}-o_t|<10^{-9}\|M\|_F^2$, where $o_t$ is the objective computed at the $t$th iteration. Under this criterion, the number of iterations for convergence is below 500 in most cases and seldom exceeds 1000. Thus, we consistently set the maximum number of iterations to 1000 in all the experiments.
\subsection{Clustering Procedure}\label{sec:main:clustering}
Given an estimate (denoted as $\hat{V}_0$) to the authentic row space $V_0$, it is rather standard to obtain the final clustering results by using $|\hat{V}_0\hat{V}_0^T|$ as an affinity matrix for spectral clustering. This approach often leads to superior clustering results, but it is time consuming especially when $n$ and $k$ are both large. For high efficiency, we shall adopt a simple and efficient approach for obtaining the final clustering results based on the estimated row space, $\hat{V}_0$, which is just an $n\times{}r$ ($r\ll{}n$) matrix.

Our approach is motivated by the following analyses. When the subspaces are independent and sufficient samples are observed for each subspace, it is known that $V_0V_0^T$ is block-diagonal and can lead to correct clustering results~\cite{ijcv_1998_factor,tpami_2013_lrr,yuxiang:nips:2013}. In this case, actually, the $n\times{}r_0$ matrix $V_0$ also owns a structure of block-diagonal. To see why, assume without loss of generality that $L_0 = [L_1,L_2,\cdots,L_k]$, where $L_i$ with SVD $U_i\Sigma_iV_i^T$ is a matrix that stores the samples from the $i$th subspace. With these notations, it is easy to see that $V_0$ is equivalent to a block-diagonal matrix; namely, $V_0=\tilde{V}_0B$ with $B\in\mathbb{R}^{r_0\times{r_0}}$ being an orthogonal matrix (i.e., $BB^T=B^TB=\Id$) and
\begin{eqnarray*}
\tilde{V}_0=\left[\begin{array}{cccc}
V_1&0&0&0\\
0&V_2&0&0\\
0&0&\ddots&0\\
0&0&0&V_k\\
\end{array}\right].
\end{eqnarray*}
Given $\tilde{V}_0$ as above, correct clustering results could be obtained by using directly the K-Means algorithm to segment the rows of $\tilde{V}_0$ into $k$ groups. Also, note that the orthogonal matrix $B$ on the right strictly preserves the inner products among the row vectors. Thus, the clustering results are the same while using $V_0$ instead of $\tilde{V}_0$ as inputs to K-Means.

The above analyses illustrate that it is appropriate to get the final clustering results by applying directly K-Means onto the row vectors of $\hat{V}_0$. Algorithm~\ref{alg:clustering} presents the whole procedure of the proposed subspace clustering method. Roughly speaking, this algorithm still falls into the category of spectral-type methods such as SIM, LRR and SSC, which obtain the final subspace clustering results by performing K-Means on a collection of top eigenvectors obtained from the eigenvalue decomposition of some matrix---for example the variants of the self-representation matrices in LRR and SSC. The main difference is that Algorithm~\ref{alg:clustering} discards the construction of $|\hat{V}_0\hat{V}_0^T|$ as well as its eigenvalue decomposition so as to produce directly the clustering results based on $\hat{V}_0$, achieving high computational efficiency.
\begin{algorithm}[htb]
\caption{Compressive Subspace Clustering by RSP}
\label{alg:clustering}
\begin{algorithmic}[1]
\STATE \textbf{input}: $M$, $R$ and $k$.
\STATE \textbf{parameters}: $r$ and $\lambda$.
\STATE \textbf{Output}: clustering results
\STATE obtain $\hat{V}_0$ by Algorithm~\ref{alg:ap}, using $M, R, r$ and $\lambda$ as the inputs.
\STATE segment the row vectors of the $n\times{}r$ matrix $\hat{V}_0$ into $k$ clusters by K-Means.
\end{algorithmic}
\end{algorithm}

\subsection{Computational Complexity}\label{sec:main:complexity}
After obtaining $\hat{V}_0$, it is known that the K-Means clustering step needs $O(nkr)$ time. So, it remains to make clear the computational complexity of Algorithm~\ref{alg:clustering}, in which the computational resources are mostly consumed by its Step 4, i.e., Algorithm~\ref{alg:ap}, which iteratively solves the non-convex optimization problem in~\eqref{eq:rsp:noisy}.

Regarding Algorithm~\ref{alg:ap}, its Step 6---the computation of the matrix $M-RS$, needs $pn+mpn$ elementary operations. The update of the variable $V$ can be finished by computing the partial-$r$ SVD of a $p\times{}n$ matrix and thus has a complexity of $O(pnr)$. To compute the gradient given in~\eqref{eq:gradient}, $mnp+4pnr+4pn$ elementary operations are required. That is, Step 8, which is indeed the most expensive step in Algorithm~\ref{alg:ap}, has an $O(mnp)$ complexity. The shrinkage operator used in Step 9 is computationally cheap, as it needs only $O(mn)$ time. In summary, each iteration in Algorithm~\ref{alg:ap} has an $O(mnp)$ complexity, thereby the overall complexity of our Algorithm~\ref{alg:clustering} is $O(mnpl+nkr)$, where $l$ is the number of iterations required by Algorithm~\ref{alg:ap} to get converged.

Up to present, the convergence rate of the alternating proximal method has not been fully understood. Empirically, we have found that at most 1000 iterations are needed for Algorithm~\ref{alg:ap} to produce near optimal solutions. So, it would be adequate to consider the computational complexity of our RPS as $O(mnp)$. Moreover, since the matrix multiplication operators are easily parallelizable, the proposed algorithms are indeed fairly fast, especially when running on Graphics Processing Unit (GPU).
\begin{figure}
\begin{center}
\subfigure[EssFace]{\includegraphics[width=0.49\textwidth]{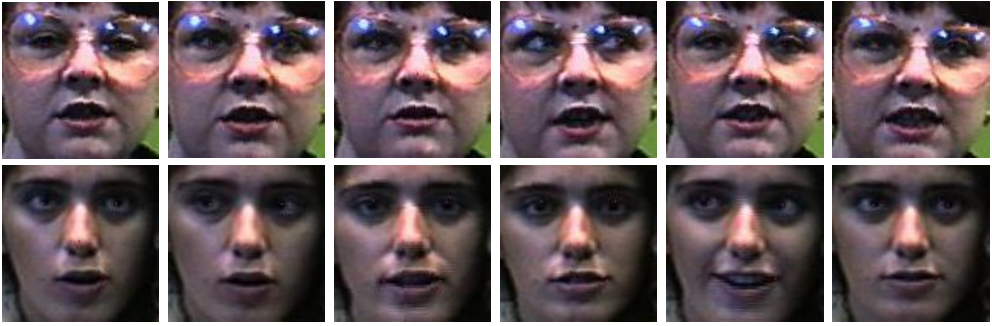}}
\subfigure[SoFace]{\includegraphics[width=0.49\textwidth]{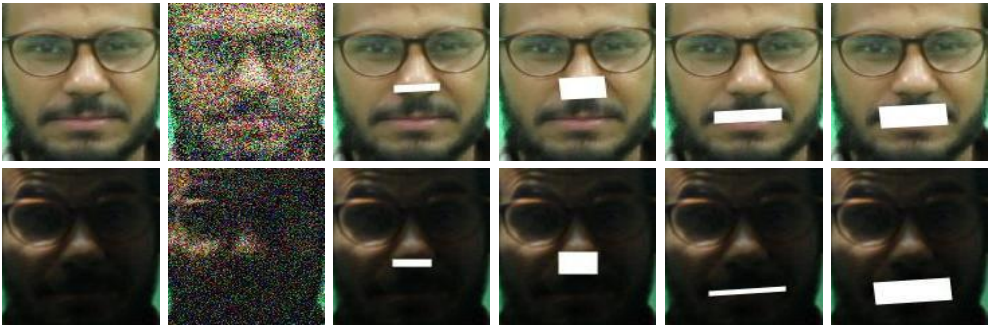}}
\subfigure[WalVideo]{\includegraphics[width=0.49\textwidth]{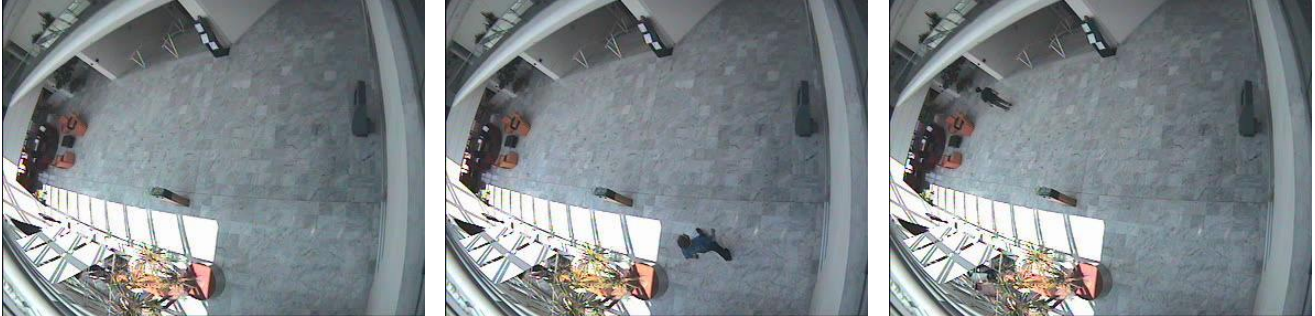}}
\vspace{-0.2in}\caption{Examples from the image datasets used in our experiments. The face or digit images in the same row across belong to the same class.}\label{fig:face:example}\vspace{-0.2in}
\end{center}
\end{figure}

Among the other things, it is worth noting that the iteration number $l$ would depend on the values of $\lambda$ and $r/p$. In general, larger $\lambda$ leads to more information loss and, accordingly, $l$ will be smaller. For example, while $\lambda$ is sufficiently large (e.g., $\lambda=+\infty$), the objective in~\eqref{eq:rsp:noisy} is perfectly minimized by $\hat{S}_0=0$, and in this case Algorithm~\ref{alg:ap} converges in only one iteration. Moreover, the iteration number $l$ also depends on the value of $r/p$. In the extreme case of $r/p\geq1$, Algorithm~\ref{alg:ap} runs only one iteration and outputs the solution of $\hat{S}_0=0$. Whenever $\hat{S}_0=0$, our RSP method is almost equivalent to applying directly SIM~\cite{ijcv_1998_factor} onto the compressed data matrix $M$.

\section{Experiments and Results}\label{sec:exp}
All experiments are conducted on a server equipped with a 64-bit Ubuntu 16.04 operating system, two Intel(R) Xeon(R) E5-2620 v4 2.10GHz CPU processors, 256GB RAM and four NVIDIA Titan X (Pascal) 12GB graphics cards. We have not implemented the algorithms using multiple GPU devices, and thus only one GPU card is randomly chosen by Matlab for accelerating the computations.
\subsection{Experimental Settings}
\begin{table}
\caption{Information about the four datasets used in the experiments of this paper .}\label{tb:expdata}\vspace{-0.2in}
\begin{center}
\begin{tabular}{|c|c|c|c|c|}\hline
     &  \#class  &\#dimension &\#points &\#points in\\
name    &  ($k$) & ($m$) & ($n$)&each class\\\hline
SynMat &2  &200 &200 &100\\
EssFace &375&10000&7495& 19$\sim$20 \\
SoFace &2662 &10000&26619&8$\sim$11\\
WalVideo &n/a&27648&1379&n/a\\\hline
\end{tabular}\vspace{-0.2in}
\end{center}
\end{table}

\subsubsection{Experimental Data}
Notice that the commonly used datasets (e.g., Hopkins155~\cite{hopkin155}) have only a few hundreds dimensions, and thus they would not be suitable for being used to verify the merits of compressive robust subspace clustering methods. As a consequence, we create four datasets for experiments, including ``SynMat'', ``EssFace'', ``SoFace'' and ``WalVideo''.

\textbf{1) SynMat:} We first consider randomly generated matrices. A collection of $200\times200$ data matrices are generated according to the model of $X=L_0+S_0$, in which $L_0$ is created by sampling 100 points from each of 2 randomly generated subspaces, the values in each point are normalized such that the super norm of $L_0$ is 1, and $S_0$ is consisting of random Bernoulli $\pm1$ values. The dimension of each subspace varies from 1 to 20 with step size 1, and thus the rank of $L_0$ varies from 2 to 40 with step size 2. The corruption size $\|S_0\|_0/n$ varies from 0.4 to 8 with step size 0.4. In summary, this dataset contains in total 400 matrices with size $200\times200$.

\textbf{2) EssFace:} The images of the second dataset we used are provided by the University of Essex\footnote{Available at cswww.essex.ac.uk/mv/allfaces/index.html.}, so referred to as ``EssFace''. This dataset contains in total 7495 images for 375 individuals, each of which has 19 or 20 images. The original images contain background, and no ground truth rectangle is provided. Thus, we utilize the face detector established in~\cite{Zhang2016Joint} to obtain the bounding boxes that contain only the faces. Then we resize the face rectangles into $100\times100$, resulting in a collection of 7495 10,000-dimensional points for experiments. Figure~\ref{fig:face:example}(a) shows some examples selected from this dataset.

\textbf{3) SoFace:} The original SoF~\cite{sof} dataset is a collection of 42,592 face images for 112 individuals, with each individual being involved in multiple photography sessions. The same physical setup is used in each session. In general, this dataset presents several challenges regarding face recognition, e.g., heavy noise, gross occlusion, strong expression, serious blurring and harsh illumination. Since the images for the same individual vary greatly in pose and appearance, it is hard, if not impossible, to form individual-level classes by using the pixel values as inputs for clustering. Thus, instead of identifying the individuals, we aim to group together the images from the same session, i.e., each session is treated as a class. Moreover, we resize the face rectangles into $100\times100$ and discard the images contaminated by blurring or canvas. For the ease of reference, we shall refer to this new version as ``SoFace'', which defines a task of segmenting 26,619 data points with dimension 10,000 into 2662 classes. Some example images from this dataset are shown in Figure~\ref{fig:face:example}(b).

\textbf{4) WalVideo:} In practice, the errors encoded in the sparse component $S_0$ could correspond to the objects of interest. To show this, we consider a surveillance video selected from the CAVIAR project\footnote{Available at homepages.inf.ed.ac.uk/rbf/CAVIAR/.}. The video we considered is a sequence of 1379 frames taken in the entrance lobby of the INRIA Labs, recording the scenes in which one person is walking in straight line, so referred to as ``WalVideo'' (see Figure~\ref{fig:face:example}(c)). This video has a near static background but contains dramatic illuminations. The original frames have a resolution $384\times288$. We reduce the resolution by half so as to obtain a $27,648\times1379$ data matrix for experiments.

For the ease of reading, we also summarize in Table~\ref{tb:expdata} the major information of the above four datasets.
\begin{figure}
\begin{center}
\includegraphics[width=0.49\textwidth]{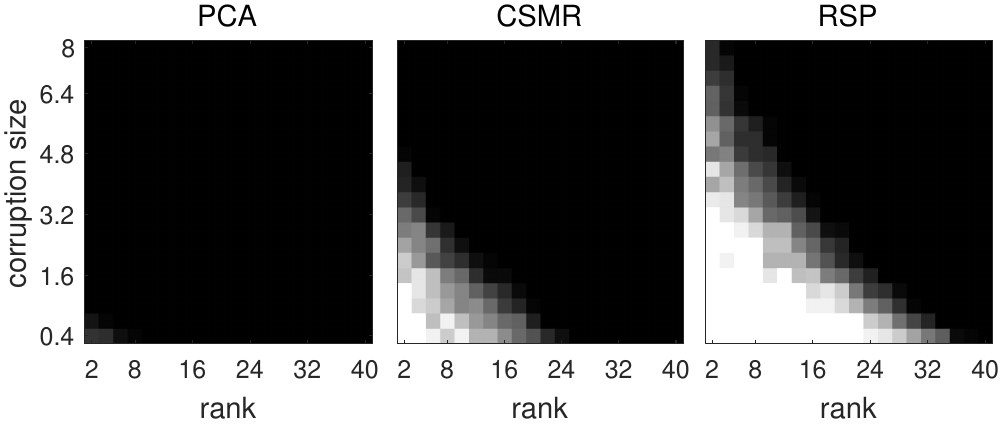}
\vspace{-0.25in}\caption{Results in recovering the randomly generated matrices in SynMat. All the methods are performed on the compressed data matrix $M$ with $p=50$. The rank $r_0$ is assumed to be given. The numbers plotted in the above figures are averaged form 20 random trials.}\label{fig:res:syn}\vspace{-0.2in}
\end{center}
\end{figure}

\subsubsection{Baselines and Evaluation Metrics}
\begin{figure}
\begin{center}
\includegraphics[width=0.49\textwidth]{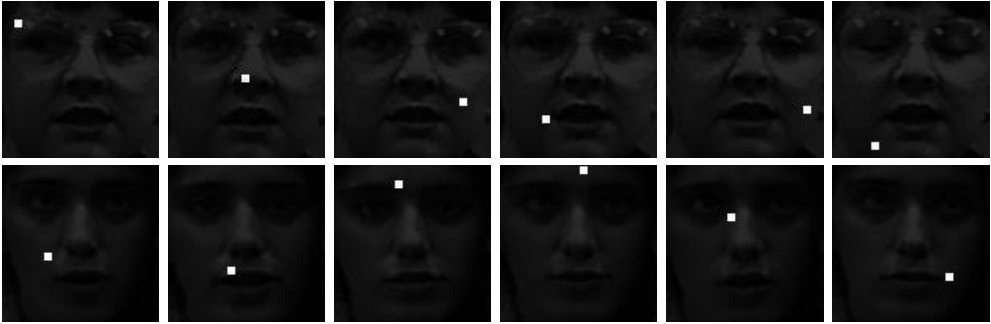}
\vspace{-0.25in}\caption{Examples from the corrupted version of EssFace, with the corruption size being chosen as 25.}\label{fig:ess:corrupt}\vspace{-0.15in}
\end{center}
\end{figure}
For the sake of comparison, we implement 6 competing methods as follows. Following the suggestions in~\cite{mao:2014:compressed,ruta:2012:compressed}, we apply three prevalent subspace clustering methods, SIM, LRR and SSC, onto the compressed data matrix $M\in\mathbb{R}^{p\times{}n}$, resulting in three benchmark baselines. Furthermore, we utilize CSMR~\cite{mardani:2013:tit} to recover $RL_0\in\mathbb{R}^{p\times{}n}$ from $M$ at first, then apply SIM, LRR and SSC onto the recovered matrices (which are estimates to $RL_0$), and in this way we obtain another three competing methods, ``CSMR+SIM'', ``CSMR+SSC'' and ``CSMR+LRR''.

Running time and clustering accuracy are used to evaluate the efficiency and effectiveness of subspace clustering methods, respectively. Here, the clustering accuracy is simply the percentage of correctly grouped data points. Also, notice that all the considered methods can be split into two stages: a learning stage that estimates some eigenvector matrix $V\in\mathbb{R}^{n\times{}l}$ ($l=k$ or $l=r$) from data, and a clustering stage that produces the final results by K-Means. So we will report their time consumption separately.
\subsubsection{Parameter Configurations}
For the ease of choosing the parameters of various subspace clustering methods, first of all, we normalize the input matrix $M$ to be column-wisely unit-normed. The parameter $r$ in SIM plays the same role as in our RSP. So, first we manually tune $r$ to maximize the accuracy of SIM, then we adjust $r$ around this estimate for RSP. The parameter $\lambda$ in RSP is chosen from the range of $2^{-10}\leq\lambda\leq2^0$. Regarding CSMR, which is indeed sensitive to its regularization parameter $\lambda$, we try our best to test as more candidates as possible from the range $2^{-10}\|R\|/\sqrt{\max(p,n)}\leq\lambda\leq2^{10}\|R\|/\sqrt{\max(p,n)}$, with the target of maximizing the accuracy of CSMR+SIM. Then the same parameter is used by the other CSMR based methods, e.g., CSMR+LRR. About the key parameter $\lambda$ in LRR and SSC, we manually select a good estimate from the range of $0.1/\sqrt{\log{}n}\leq\lambda\leq10/\sqrt{\log{}n}$ and $2^{-10}\leq\lambda\leq2^{10}$, respectively.
\begin{figure}
\begin{center}
\includegraphics[width=0.49\textwidth]{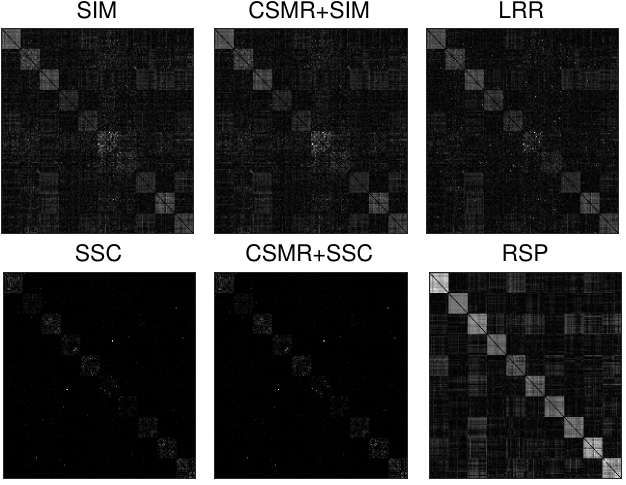}
\vspace{-0.25in}\caption{Visualization of the affinities matrices corresponding to 10 classes in EssFace. For RSP that needs only $\hat{V}_0$, we show $\hat{V}_0\hat{V}_0^T$ for the sake of comparison. Each affinity matrix is post-processed by zeroing-out its diagonal and normalizing its values to have a maximum of 1.}\label{fig:ess:affinity}\vspace{-0.2in}
\end{center}
\end{figure}
\subsection{Results on SynMat}
\begin{table}
\caption{Evaluation results on the corrupted version of EssFace, with corruption size 25 and $p=500$. Our GPU is of single precision, thereby the accuracy achieved using GPU is slightly lower than using CPU.}\label{tb:res:ess}\vspace{-0.15in}
\begin{center}
\begin{tabular}{cc|ccc}\hline
method    &  accuracy (\%) & \multicolumn{3}{c}{time (seconds)}\\\cline{3-5}
       &                   & learning & k-means & total\\\hline
SIM    &  12.25            &\textbf{24}    &43 & \textbf{67} \\
CSMR+SIM  &  12.84         &499    &42 & 541\\
LRR       &  7.45          &105    &37 & 142\\
CSMR+LRR  &  7.33          &637    &37 &674\\
SSC       &  9.14          &1439   &44 &1483\\
CSMR+SSC  &  9.19          &2903  &45 &2948\\\hline
RSP (CPU) & \textbf{66.76} &1177   &\textbf{20} &1197\\
RSP (GPU) &  66.71         &257      &23 &280\\\hline
\end{tabular}
\end{center}\vspace{-0.2in}
\end{table}
As mentioned in Section~\ref{sec:main:rsp}, it is possible that RSP strictly succeeds in recovering $V_0V_0^T$ under certain conditions. To verify this, we first experiment with the SynMat dataset. The $200\times200$ matrices are projected to $50\times200$ by RP with $p=50$, and task here is to recover the authentic row space by using only the compressed matrix. For each pair of $r_0$ and $\|S_0\|_0/n$, we perform 20 random trials, and thus in this experiment we run 8000 simulations in total. To show the superiorities of RSP, we also consider to recover the target $V_0V_0^T$ by PCA and CSMR: PCA estimates $V_0V_0^T$ by computing directly the SVD of $M$, while CSMR is to firstly obtain an estimate to $RL_0$ by program~\eqref{eq:csmr} and then try obtaining $V_0V_0^T$ via performing SVD on the estimate. The accuracy of recovery, i.e., the similarity between $V_0V_0^T$ and $\hat{V}_0\hat{V}_0^T$, is measured by Signal-to-Noise Ratio, $\mathrm{SNR_{dB}}$.

The evaluation results are shown in Figure~\ref{fig:res:syn}, in which each plotted number is a score defined as in the following:
\begin{align}\label{eq:score}
\mathrm{score} = \left\{\begin{array}{cc}
0, & \mathrm{SNR_{dB}}<15,\\
0.2, & 15\leq\mathrm{SNR_{dB}}<20,\\
0.5, & 20\leq\mathrm{SNR_{dB}}<30,\\
1,   & \mathrm{SNR_{dB}}\geq30.
\end{array}\right.
\end{align}
As we can see, PCA works poorly, attaining $\mathrm{SNR_{dB}}$ smaller than 15 in almost all the cases. This illustrates that it is unlikely to solve Problem~\ref{pb:crsr} without accessing the sensing matrix $R$. Also, it can be seen that CSMR (with $\lambda=1.2\|R\|/\sqrt{\max(p,n)}$) succeeds only in limited cases. The reason is that, as aforementioned, CSMR essentially requires $r_0\ll\min(p,n)$ such that $RL_0$ is low rank. Our RSP may partially overcome this limit, thereby RSP (with $r=r_0$ and $\lambda=2^{-7}$) can do much better than CSMR in recovering the authentic row space.
\begin{figure}
\begin{center}
\includegraphics[width=0.49\textwidth]{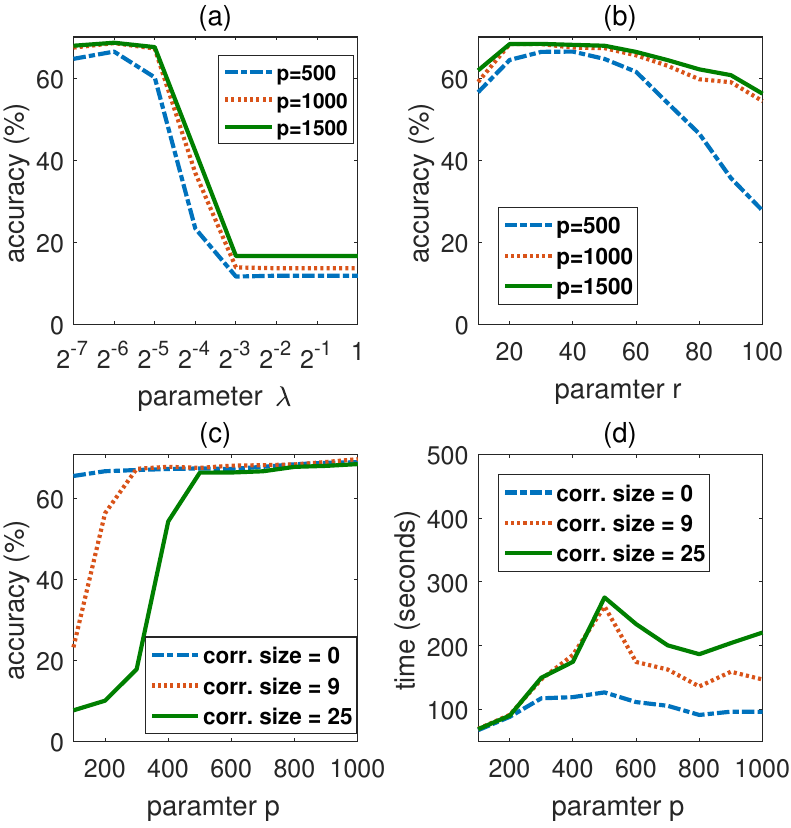}
\vspace{-0.25in}\caption{Explore the performance of RSP under various parametric settings, using EssFace as the experimental data. (a) The parameter $\lambda$ is varying while $r=40$ and corruption size $=25$. (b) The parameter $r$ is varying while $\lambda=2^{-6}$ and corruption size $=25$. (c-d) The parameter $p$ is varying while $\lambda=2^{-6}$ and $r=40$.}\label{fig:ess:para}\vspace{-0.2in}
\end{center}
\end{figure}

\subsection{Results on EssFace}
To get a comprehensive understanding about RSP, we corrupt each image by adding an $s\times{}s$ ($s=0,3,5$) spot at a location randomly chosen from the image rectangle. The values in the spot are made to be 5 times as large as the maximum of pixel values so as to suppress the visual information of faces (see Figure~\ref{fig:ess:corrupt}).
\begin{table}
\caption{Evaluation results on SoFace, with $p=500$.}\label{tb:res:sof}\vspace{-0.15in}
\begin{center}
\begin{tabular}{cc|ccc}\hline
method    &  accuracy (\%) & \multicolumn{3}{c}{time (seconds)}\\\cline{3-5}
       &                   & learning & k-means & total\\\hline
SIM           &  52.16         &1801&4005&5806\\
CSMR+SIM      &  53.82         &2809&3998& 6807\\
LRR           &  48.18         &2984&4012& 6996\\
CSMR+LRR      &  48.01         &4247&4025& 8272\\
SSC           & 53.44          &20777&4055&24832\\
CSMR+SSC      & 53.79          &23070&4087&27157\\\hline
RSP (CPU)     &\textbf{57.61}  & 1315&  \textbf{310} &1625\\
RSP (GPU)     &   57.42        &\textbf{214}&   314&\textbf{528}\\\hline
\end{tabular}\vspace{-0.2in}
\end{center}
\end{table}

Table~\ref{tb:res:ess} and Figure~\ref{fig:ess:affinity} show the comparison results at $s=5$ (i.e., the corruption size is 25) and $p=500$. It can be seen that all of SIM, LRR and SSC produce very poor results. This is because these methods possess no mechanism to disentangle the authentic samples and gross errors. In fact, given only the compressed data $M$, there is no way to correctly segment this dataset. The pre-processing of CSMR fails to make any substantial improvement in terms of clustering accuracy. The reason is that $p=500$ is not large enough for CSMR to get a good estimate to $RL_0$. In contrast, our RSP (with $r=40$ and $\lambda=2^{-6}$) can achieve an accuracy about 67\%. This result, in fact, is even slightly better than the best result that SIM, LRR and SSC achieved on the original 10000-dimensional data: Using PCP to pre-process the original data matrix $X$, SIM, LRR and SSC attain accuracies of 64.23\%, 61.32\% and 65.19\%, respectively. However, it is very time-consuming to perform robust subspace clustering on the original high-dimensional data. Namely, PCP spends more than 23 hours in recovering $L_0$. In sharp contrast, our RSP (uisng GPU) needs only about 5 minutes to get good clustering results with accuracy about 67\%.
\begin{figure*}
\begin{center}
\includegraphics[width=0.95\textwidth]{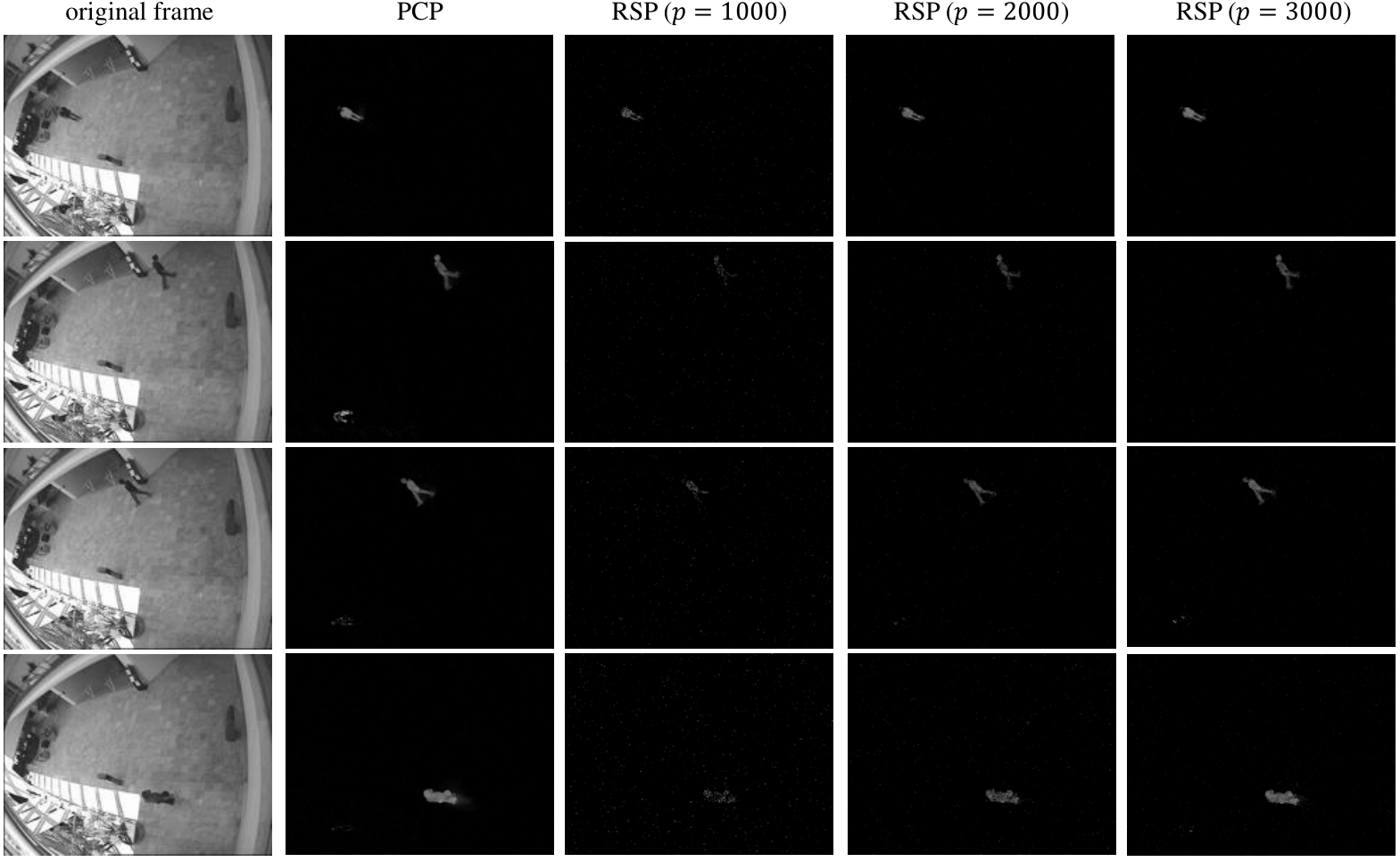}
\vspace{-0.1in}\caption{Moving object detection in surveillance video. Four frames from the WalVideo dataset, with varying illumination. From left to right: the original frame, the sparse component $|\hat{S}_0|$ obtained by PCP and RSP. The parameters in RSP are set as $r=5$ and $\lambda=2^{-6}$.}\label{fig:wal:res}\vspace{-0.2in}
\end{center}
\end{figure*}

We also investigate the influences of the parameters in RSP as well as the projection number $p$. As shown in Figure~\ref{fig:ess:para}(a), the accuracy of RSP drops dramatically when $\lambda\geq2^{-4}$. This is because, as aforementioned, RSP will converge to the trivial solution $\hat{S}_0=0$ if $\lambda$ is sufficiently large. Provided that there is no dense noise in the data, theoretically speaking, there exists $\lambda^*>0$ such that RSP works equally well for all $\lambda\leq\lambda^*$. However, in practice the white noise is ubiquitous, and thus the performance of RSP slightly degrades while $\lambda$ is too small. Overall, $\lambda=2^{-6}$ is a good choice for this dataset. Regarding the parameter $r$, Figure~\ref{fig:ess:para}(b) shows that RSP could work almost equally well while $r$ locates in a certain range. This confirms our doctrine that $r$ is unnecessary to be identical to $r_0$. For this dataset, $r=40$ is a proper setting, as can be seen from Figure~\ref{fig:ess:para}(b). As we can see from Figure~\ref{fig:ess:para}(c), RSP breaks down while $p$ is too small, and the value of the breaking point depends on the corruption size. More precisely, without the gross corruptions, RSP actually works equally well for a wide range of $p$. In the case where the corruption size is 9, RSP breaks down when $p\leq200$. When the corruption size increases to 25, the breaking point becomes $p\leq400$. These phenomena, in general, are consistent with the conditions listed in~\eqref{eq:condition}. Figure~\ref{fig:ess:para}(d) plots the running time as a function of the parameter $p$, revealing the phenomenon that the running time of RSP does not increase monotonically with $p$. This is because, as discussed earlier, the convergence speed of Algorithm~\ref{alg:ap} actually depends on $p$.

\subsection{Results on SoFace}
To verify the effectiveness of various methods under the context of compressive robust subspace clustering, we reduce the data dimension to 500 by RP. The comparison results are shown in Table~\ref{tb:res:sof}. In terms of running time, RSP (with $r=120$ and $\lambda=2^{-5}$) distinctly outperforms all the competing methods. In particular, RSP is even much faster than SIM, which is to simply apply SIM onto the compressed matrix $M\in\mathbb{R}^{500\times26619}$. This is because SoFace has a large number of data points and classes, saying $n=26619$ and $k=2662$. In this case, spectral clustering is indeed very time-consuming due to the following two procedures: 1) computing the partial SVD of an $n\times{}n$ matrix, and 2) using K-Means to segment a collection of $n$ $k$-dimensional points into $k$ clusters. On this dataset, our Algorithm~\ref{alg:ap} converges with about 400 iterations, and after that, in sharp contrast, Algorithm~\ref{alg:clustering} only needs to perform K-Means clustering on a set of $n$ $r$-dimensional points. Besides of its high computational efficiency, RSP also outperforms the competing methods in the sense of clustering accuracy.
\begin{table}
\caption{Running time on WalVideo. The parameters of RSP are set as $r=5$ and $\lambda=2^{-6}$}\label{tb:res:wal}\vspace{-0.15in}
\begin{center}
\begin{tabular}{ccc}\hline
          &   \multicolumn{2}{c}{time (seconds) }\\
method    &  CPU      & GPU\\\hline
PCP       &   4606        &2820 \\\hline
RSP ($p=1000$)& 1024      &301 \\
RSP ($p=2000$)& 1755      &469 \\
RSP ($p=3000$)& 2299      &511 \\\hline
\end{tabular}\vspace{-0.2in}
\end{center}
\end{table}
\subsection{Results on WalVideo}
Unlike the above clustering experiments, in this experiment the input matrix $M$ is not normalized. This is because we need to visualize the sparse component $\hat{S}_0$ produced by RSP. Figure~\ref{fig:wal:res} shows four frames taken from the WalVideo dataset, which has dramatic illuminations in background. As we can see, RSP with $p\geq2000$ works as well as PCP. What is more, in terms of stability against the illumination in background, RSP is even slightly better than PCP. To be more precise, PCP occasionally treats a considerable amount of background illumination as the moving objects (see the second row of Figure~\ref{fig:wal:res}), while RSP produces more reliable results for the same frame. Since in this dataset the data matrix $X$ is tall (i.e., $m\gg{}n$), the computational complexity of RSP and PCP has the same order. Yet, as shown in Table~\ref{tb:res:wal}, RSP is still faster than PCP, especially when GPU is used. In particular, our RSP is more parallelizable than PCP, and thus RSP benefits more from GPU than PCP does, as we can see from Table~\ref{tb:res:wal}.
\section{Conclusion}
In this paper we studied the problem of \emph{compressive robust subspace clustering}, a significant problem not thoughtfully explored before. We first mathematically formulated the problem as to recover the row space of the clean data, given only the compressed data $M$ and sensing matrix $R$. Then we devised a simple method termed RSP, which iteratively seeks both the authentic row space and the sparse errors possibly existing in the original high-dimensional data. Extensive experiments with various settings verified the effectiveness and efficiency of RSP.

There are still several problem left for future work. For example, it is better to estimate or learn the hyper-parameter $r$ from the data. It is also of considerable significant to explore the tensor form and multi-view extensions of the proposed problem.
\section*{Acknowledgement}
Thanks Dr. Yubao Sun and Dr. Meng Wang for sharing with us their opinions. This work is supported in part by national Natural Science Foundation of China (NSFC) under Grant 61622305, Grant 61825601, Grant 61532009, Grant 61672365, Grant 61425011, Grant 61720106001 and Grant 61529101, in part by Natural Science Foundation of Jiangsu Province of China (NSFJPC) under Grant BK20160040, in part by SenseTime Research Fund.
\bibliographystyle{IEEEtran}
\bibliography{ref}
\end{document}